\documentclass[journal]{IEEEtran}

\usepackage{epsfig}
\usepackage{graphicx}
\usepackage{amsmath}
\usepackage{amssymb}
\usepackage{graphics} 
\usepackage{mathptmx} 
\usepackage{booktabs}
\usepackage{multirow}
\usepackage{color}
\usepackage{mathrsfs}
\usepackage{float}
\usepackage{subfigure}
\usepackage{url}

\ifCLASSOPTIONcompsoc
  \usepackage[nocompress]{cite}
\else
  \usepackage{cite}
\fi

\ifCLASSINFOpdf
\else
\fi
\begin{document}
%
\title{\textbf{Masked GAN for Unsupervised Depth and Pose Prediction with Scale Consistency}}
%
%
%

\author{Chaoqiang~Zhao,
        Gary G. Yen,~\IEEEmembership{Fellow,~IEEE,}
		Qiyu~Sun,
        Chongzhen~Zhang,
		Yang~Tang,~\IEEEmembership{Senior Member,~IEEE}
\IEEEcompsocitemizethanks{\IEEEcompsocthanksitem This work was supported in part by the National Natural Science Foundation of China under Grant Nos. 61988101, 61673176, in part by the Program of Shanghai Academic Research Leader under Grant No. 20XD1401300, in part by the Programme of Introducing Talents of Discipline to Universities (the 111 Project) under Grant B17017 (\textit{Corresponding author: Yang Tang}).
\IEEEcompsocthanksitem C. Zhao, Q. Sun, C. Zhang and Y. Tang are with the Key Laboratory of Advanced Control and Optimization for Chemical Process, Ministry of Education, East China University of Science and Technology, Shanghai, 200237, China (e-mail: zhaocq@mail.ecust.edu.cn (C. Zhao); yangtang@ecust.edu.cn, tangtany@gmail.com (Y. Tang)).
\IEEEcompsocthanksitem Gary G. Yen is with the School of Electrical and Computer Engineering, Oklahoma State University, Stillwater, OK 74075, USA (e-mail: gyen@ okstate.edu).
}
}

%
%

\markboth{}%
{Shell \MakeLowercase{\textit{et al.}}: Bare Demo of IEEEtran.cls for IEEE Journals}
%



\maketitle

\begin{abstract}
Previous work has shown that adversarial learning can be used for unsupervised monocular depth and visual odometry (VO) estimation, in which the adversarial loss and the geometric image reconstruction loss are utilized as the mainly supervisory signals to train the whole unsupervised framework.
However, the performance of the adversarial framework and image reconstruction is usually limited by occlusions and the visual
field changes between frames.
This paper proposes a masked generative adversarial network (GAN) for unsupervised monocular depth and ego-motion estimation.
The MaskNet and Boolean mask scheme are designed in this framework to eliminate the effects of occlusions and impacts of visual field changes on the reconstruction loss and adversarial loss, respectively.
Furthermore, we also consider the scale consistency of our pose network by utilizing a new scale-consistency loss, and therefore, our pose network is capable of providing the full camera trajectory over a long monocular sequence.
Extensive experiments on the KITTI dataset show that each component proposed in this paper contributes to the performance, and both our depth and trajectory predictions achieve competitive performance on the KITTI and Make3D datasets.
\end{abstract}

\begin{IEEEkeywords}
Adversarial learning, unsupervised learning, depth estimation, visual odometry, GAN, scale consistency.
\end{IEEEkeywords}

%
\IEEEpeerreviewmaketitle

\section{Introduction}

Understanding the 3D structure of a scene and estimating the ego-motion are two basic tasks of autonomous robots \cite{eigen2014depth,zhou2017unsupervised}. With the development in deep learning (DL) technology, DL-based depth and pose prediction have achieved outstanding results in both supervised \cite{eigen2014depth} and unsupervised manners \cite{zhou2017unsupervised}. Because they are free from expensive training methods that involve ground truth during training, unsupervised methods have been widely studied \cite{zhao2020monocular,tang2020perception}, in which depth and pose networks are jointly trained by monocular videos.
The principle is that one can warp the image in one frame (source frame) to another frame (target frame) using the projection based on predicted depth and ego-motion \cite{zhou2017unsupervised}, which is called image warping or view reconstruction, and the difference between the real-world and synthesized images is regarded as the main supervisory signal instead of the ground truth during training.
However, occlusions as well as the visual field changes between frames influence the quality of the synthesized images, which will also affect the view reconstruction and the training of unsupervised frameworks.
Recent studies have combined masks \cite{zhou2017unsupervised}, semantic segmentation \cite{chen2019towards} or motion segmentation \cite{ranjan2019competitive} networks for multi-task training framework, thereby further improving the performance of the networks.

\begin{figure}[!t]
	\centering
	
	\includegraphics[width = \columnwidth]{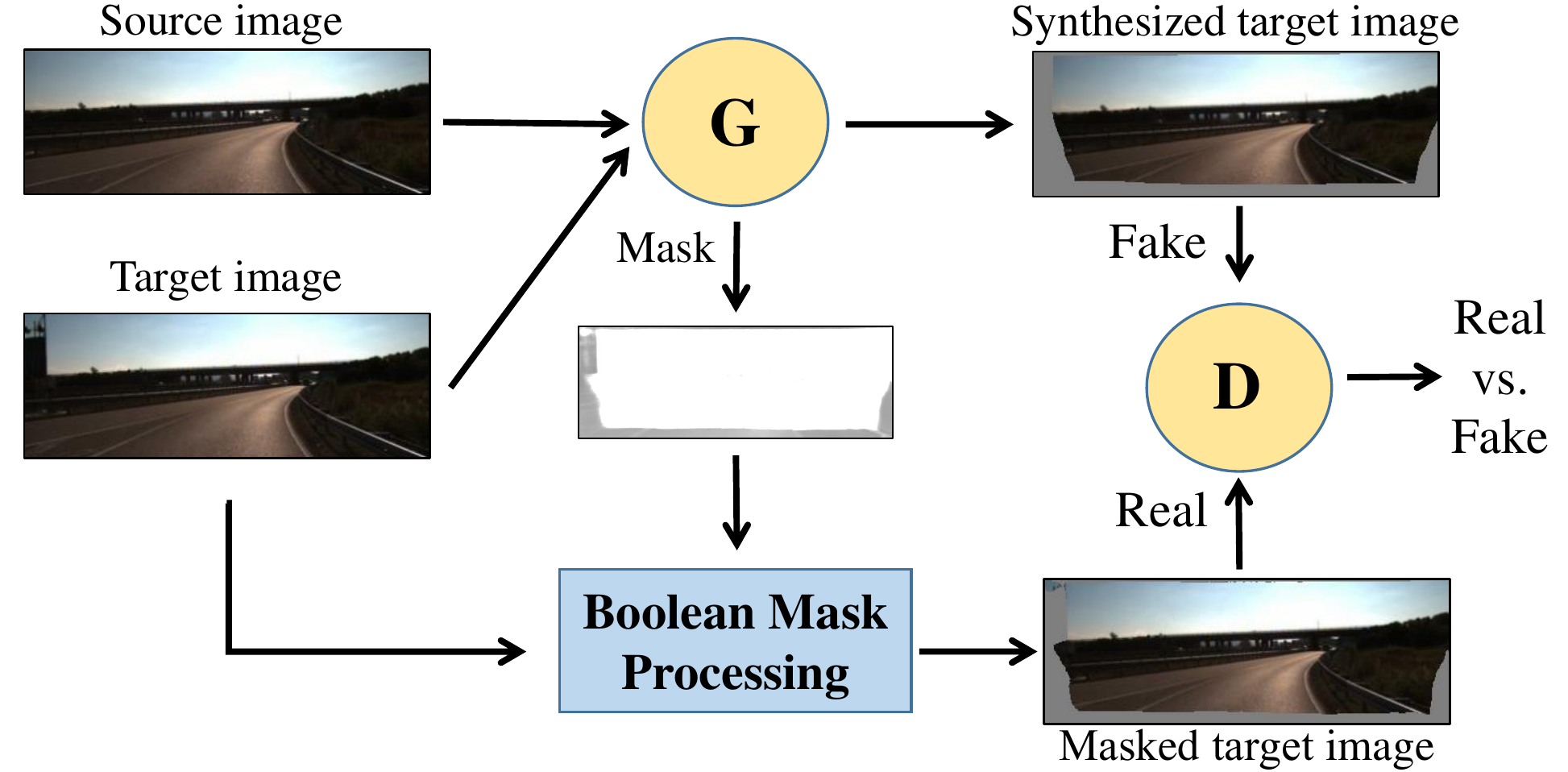}
	\caption{Architecture overview. Our unsupervised adversarial learning framework consists of two major components, generator (G) and discriminator (D). Adjacent frames are taken to G to reconstruct target image and predict a mask of unreconstructed regions. Instead of sending the target image to discriminator directly, a ``Boolean mask processing'' step is designed to preprocess it. As shown in the ``masked target image", the unreconstructed regions similar to synthesized image are created, thereby reducing the effect of unreconstructed regions on discriminator.}
	\label{fig:fig1}
\end{figure}

\begin{figure}[t]
	\centering
	\includegraphics[width = \columnwidth]{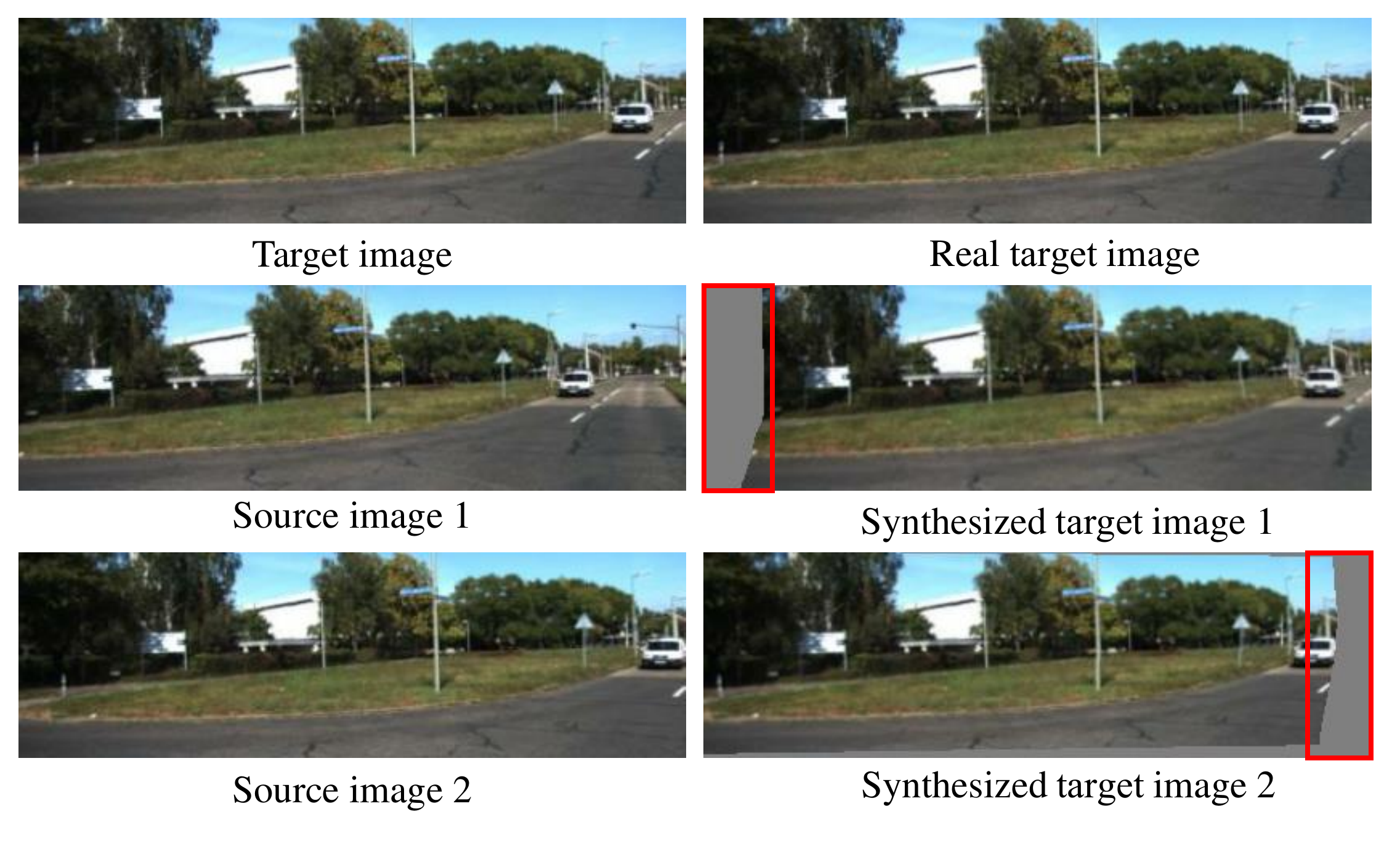}
	\caption{Comparison of the real target image and the synthesized target images. The synthesized target images (synthesized images 1 and 2) are synthesized from corresponding source images (source images 1 and 2) by view reconstruction algorithm. Due to the motion, the visual field changes between adjacent images, which causes the incomplete pixel correspondence between the target and source images. Therefore, edge regions in synthesized images cannot be synthesized from the source images because of the information miss caused by visual field changes.}
	\label{fig:fig2}
\end{figure}

Adversarial learning has shown strong capabilities in the field of image processing recently \cite{zhang2019adversarial}. Recent studies have demonstrated that introducing adversarial learning for depth and pose estimation can significantly improve the accuracy of these two tasks \cite{cs2018monocular,almalioglu2019ganvo,zhao2019geometry,jung2017depth,pilzer2018unsupervised}.
In \cite{cs2018monocular,almalioglu2019ganvo,li2019sequential}, adversarial learning is combined with the framework of unsupervised methods for depth and pose estimation based on monocular videos.
In their frameworks \cite{cs2018monocular,almalioglu2019ganvo,li2019sequential}, a discriminator is designed to distinguish between the real-world images and the synthesized images, and the min-max game between the generator and discriminator is used to improve the performance of both the depth and VO networks.
However, because of the occlusions and the visual field changes between frames, the quality of synthesized images is affected, causing the unreconstructed regions in synthesized images, which are inevitable and shown in Figs. \ref{fig:fig1} and \ref{fig:fig2}.
If these inevitable distinguishing features are learned by the discriminator, then the min-max game between the generator and discriminator will be broken and the performance of the total framework is limited, which is overlooked by previous works \cite{cs2018monocular,almalioglu2019ganvo,li2019sequential}. Finally, the performance of the generator (depth and pose networks) will be heavily affected.
In this paper, we focus on eliminating the effects of occlusions and the visual field changes between frames on adversarial loss and propose a Boolean mask processing step to improve the training of the adversarial framework. Our framework is shown in Fig. \ref{fig:fig1}. 

Compared with the traditional visual odometry (VO) or structure from motion (SfM) methods \cite{engel2017direct,mur2017orb}, DL-based methods can estimate the 6-DOF (degrees of freedom) ego-motion and pixel-level dense depth maps in an end-to-end manner without back-end optimization process.
Previous unsupervised methods, such as \cite{ zhou2017unsupervised,yin2018geonet,ranjan2019competitive,almalioglu2019ganvo}, are trained by short monocular frame snippets. There is lack of a suitable loss function to constrain the scale consistency of the predicted results among the different snippets, $i.e.$, the scale factors of the poses and depth maps predicted by the networks in two different snippets are different. As a result, because of the scale-inconsistency among different image snippets, the global trajectory of the monocular videos cannot be provided by the pose network, and the scale-inconsistent depth maps cannot be used in practice. To have a consistent scale estimation of a pose network, Bian \textit{et al.} \cite{bian2019depth} consider this challenge and propose a geometric consistency loss to align the scale factor of different depth maps. However, they only constrain the consistency of values between depth maps and ignore the consistency of structures, resulting in performance limited. Therefore, we utilize the structural similarity (SSIM) \cite{wang2004image} to further constrain the structural similarity and scale-consistency and get a better result.

Motivated by the limitation in previous studies and the outstanding performance of adversarial learning, this paper proposes a novel unsupervised adversarial framework for monocular depth and ego-motion estimation. A MaskNet and a Boolean mask scheme are proposed to eliminate the impacts of occlusions and visual field changes between frames on reconstruction loss and especially adversarial loss. In addition, a novel adaptive loss function based on SSIM is also designed in this paper to strengthen the scale consistency of our pose network.
Moreover, we conduct detailed ablation studies to clearly demonstrate the effectiveness of the proposed unsupervised adversarial framework. Comprehensive evaluation results on the KITTI\cite{geiger2013vision} and Make3D datasets \cite{saxena2008make3d} show that our proposed framework obtains a competitive accuracy and transferability. Furthermore, our pose network has the ability to provide an accurate trajectory over a long monocular video.

In summary, our main contributions are as follows:
\begin{itemize}
\item  We introduce a masked GAN framework for pose and depth estimation, where the effects of occlusion and visual field changes on view reconstruction are considered.
\item  We discuss the effect of unreconstructed regions on adversarial learning, which is ignored in previous work, and Boolean mask processing is proposed in this paper to eliminate this negative influence.
\item  We consider the scale-inconsistent problem and propose a adaptive constraint for a better global trajectory prediction. At the same time, both the pose and depth networks proposed in this paper show competitive results on public datasets.
\end{itemize}

In this paper, previous work on monocular ego-motion and depth prediction are discussed in Section II. Section III introduces the proposed unsupervised masked GAN framework in detail. Section IV shows our experimental results using the proposed method on KITTI \cite{geiger2013vision} and Make3D datasets \cite{saxena2008make3d}. Finally, this study is concluded in Section V.

\section{Related Work}

Traditional 3D structure recovery is based on the triangulation algorithm \cite{mur2017orb}, which must find a set of matching pixels between multiple frames. With the technological advancements in DL, convolutional neural networks (CNNs) have shown their superb ability in monocular depth estimation. In this section, we review the previous works on DL-based monocular depth and ego-motion estimation.

\textbf{Learning from the ground truth.} Eigen \textit{et al.} \cite{eigen2014depth} introduce CNNs into monocular depth estimation and predict the depth in an end-to-end manner. Alex \textit{et al.} \cite{kendall2015posenet} design a PoseNet based on CNNs for 6-DOF pose regression. Recently, in \cite{xue2019beyond,cs2018depthnet}, recurrent neural networks are also adopted to extract temporal features and preserve accumulated information to improve the estimation accuracy. Although the above methods \cite{xue2019beyond,cs2018depthnet} achieve satisfactory accuracy in pose and depth estimation, both of them rely on the ground truth as the supervisory signal, and the ground truth is difficult and expensive to acquire.

Recently, unsupervised monocular depth and ego-motion estimation has been well investigated because it is free from the ground truth. With regard to the training approaches, unsupervised depth learning methods can be divided into two types: learning from stereo image pairs \cite{garg2016unsupervised,godard2017unsupervised,zhan2018unsupervised} and learning from monocular videos \cite{zhou2017unsupervised,yin2018geonet,wang2019unsupervised}. Although the training process of these methods depends on the geometric constraints between multi-frame, the trained networks can predict depth maps from a single image independently during testing.

\textbf{Learning from stereo image pairs.} Garg \textit{et al.} \cite{garg2016unsupervised} prove that a depth network can be trained by stereo image pairs in an unsupervised manner. They utilize the inverse depth prediction and the geometry between the stereo image pairs to reconstruct (inverse warping) the left image from the right image. The difference between synthesized and real target images is used as a supervisory signal during training. Godard \textit{et al.} \cite{godard2017unsupervised} follow and expand this idea by using a left-right consistency constraint to achieve better performance. In addition, in \cite{zhan2018unsupervised}, authors introduce pose estimation into the framework, and the networks are jointly trained on stereo sequences. The geometric constraints of the temporal (image sequences) and spatial (stereo image pairs) aspects are utilized to improve the performance of the ego-motion and depth estimation.
Although the above method \cite{zhan2018unsupervised} can estimate the pose and depth with scale information, the accuracy relies heavily on having accurate calibration between the stereo cameras.

\textbf{Learning from monocular videos.} Considering the advantages of a single camera system, such as small size and low power consumption, unsupervised methods based on monocular sequences have been proposed to train the ego-motion and depth prediction networks. Zhou \textit{et al.} \cite{zhou2017unsupervised} propose a framework in which the depth network is jointly trained with the pose network by using monocular videos. They reconstruct the current image from its adjacent frames by view reconstruction, which relies on the output of pose and depth networks. Then, the reconstruction loss between the reconstructed and raw images is computed as supervisory signals. Afterwards, several researchers follow and extend \cite{zhou2017unsupervised} this approach into multiple tasks \cite{yin2018geonet,ranjan2019competitive,chen2019towards,zou2018df}, and the intrinsic geometric constraints among the different tasks are utilized to strengthen the supervised signal and improve the training process. For example, Ranjan \textit{et al.} \cite{ranjan2019competitive} combine four fundamental problems (depth prediction, ego-motion estimation, optical flow and motion segmentation) through geometric constraints. In addition, a competitive training approach is proposed to balance the training process of each network. Various sensor data \cite{lee2019depth,imran2019depth,chen2019selective} are also used for depth and pose estimation. Nevertheless, due to the lack of appropriate scale consistent constraints, the pose network trained by unlabelled frame snippets cannot generate the full trajectory of a long video sequence. Therefore, Bian \textit{et al.} \cite{bian2019depth} tackle this challenge by a geometric loss for scale consistency.

\textbf{Learning with generative adversarial networks (GANs).} Because of the outstanding performance of GANs on image processing, introducing adversarial learning into monocular depth estimation is becoming a hot topic. In the adversarial learning framework \cite{goodfellow2014generative}, a generator is designed to learn and mimic the distribution of real data, and a discriminator is designed to assess the quality of the generated data and promote the performance of the generator. Kumar \textit{et al.} \cite{cs2018monocular} apply GANs to monocular depth estimation. Their generator consists of depth and pose networks, and the outputs of the networks are used to reconstruct the images by view construction. At the same time, a discriminator is designed to distinguish between the synthesized and real images. Recently, Almalioglu \textit{et al.} \cite{almalioglu2019ganvo} combine a recurrent learning approach with GAN for pose and depth estimation in an unsupervised manner. They leverage a long short-term memory module (LSTM) to extract temporal information for pose estimation. At the same time, their generator estimates the depth map from a random vector $z$, i.e., the depth network cannot predict the depth map from a single image in an end-to-end manner.
The most similar work to this paper is proposed in \cite{li2019sequential}. Li \textit{et al.} introduce a mask network to reduce the effects of dynamic objects and occlusions on reconstruction loss, and an LSTM module is used for depth estimation. However, their depth network takes one single image and sequence information extracted by the LSTM as input for the depth estimation, i.e., their depth network can only be used on monocular sequences and cannot estimate the depth map from a single image, which is different from the proposed depth network herein. In addition, they do not consider the effects of occlusions on adversarial loss, which results in limited performance. In this paper, we propose a novel unsupervised adversarial framework for monocular depth and ego-motion estimation. A Boolean mask scheme is proposed to eliminate the effects of occlusions on adversarial loss, thereby getting a better performance.

Moreover, the above GAN-based methods \cite{cs2018monocular,almalioglu2019ganvo,li2019sequential} ignore the influence of occlusions and the visual field changes between adjacent frames in the adversarial learning.
The pixels between the target and source images do not correspond exactly because of the visual field changes caused by motion, and thus, the target images cannot be reconstructed completely from the source images by view reconstruction algorithms and bilinear interpolation, as shown in Fig. \ref{fig:fig2}. Therefore, the data distribution of the unreconstructed regions in the synthesized images is unique and cannot be eliminated exactly by the training of the generator. These unique distributions will be learned by the discriminator, which will affect the adversarial learning process and the performance of the generator, as shown in the experiments.

In this paper, we present a novel loss function to constrain the scale-consistency of our pose and depth networks. At the same time, considering the influence of the unreconstructed regions on the discriminator, which is overlooked in previous work, we design a mask network to estimate these regions, and we introduce Boolean mask processing to eliminate their influence.

\begin{figure*}[!htbp]
	\centering
	\includegraphics[width = 1.9\columnwidth]{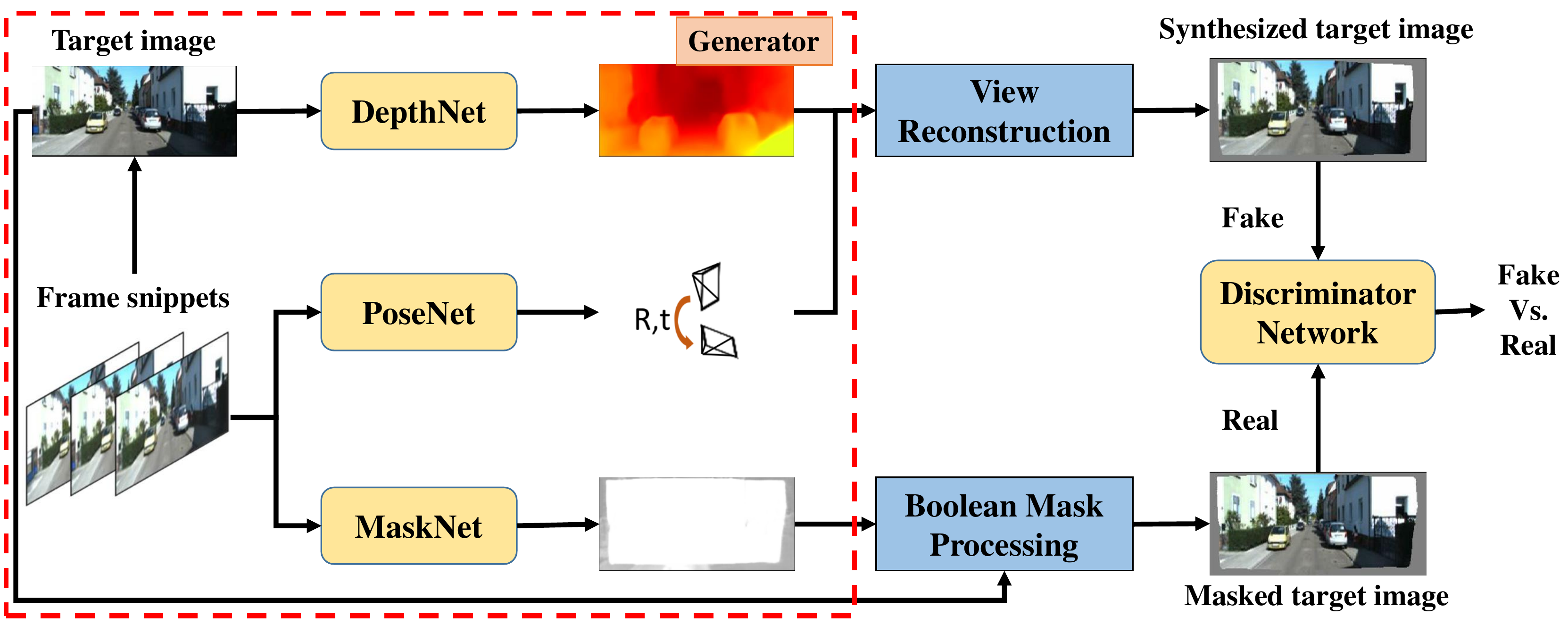}
	\caption{Our unsupervised framework consists of two major parts, a generator and a discriminator. \textbf{Generator:} The main task of the generator is to reconstruct an image with the same data distribution as the target image by using the results of PoseNet and DepthNet. Due to the factors such as occlusions and visual field changes, some regions cannot be reconstructed (shown in the synthesized target image), which will help the discriminator distinguish between real and fake images. Therefore, a MaskNet is designed to estimate these unreconstructed regions. We couple our MaskNet with PoseNet for fewer parameters and easier training, which is similar to \cite{zhou2017unsupervised}. In order to show our full architecture clearly, we draw the PoseNet and MaskNet separately here. \textbf{Discriminator:} Before sending the real and synthesized target images to discriminator, we design the Boolean mask processing step based on the output of MaskNet to process the real target image. Hence, the major difference (unreconstructed regions) between real and fake images is removed, and this will make the discriminator learn a deeper feature and data distribution.} 
	\label{fig:fig3}
\end{figure*}

\section{Methodology}

In this section, we will give a brief introduction to the network architecture proposed in this paper, the unsupervised training framework, and the overall supervisory signals.
\subsection{Architecture overview}

The framework of our unsupervised network is shown in Fig. \ref{fig:fig3}. The generator takes a short sequential frame snippet that consists of a target image $I_{t}$ and an adjacent source image $I_{s}$; in addition, the output of the generator contains a depth map, a 6-DOF pose, and the Boolean mask $M_{b}$, which corresponds to the unreconstructed region in the synthesized images.
Based on the predicted depth and pose, the view reconstruction algorithm that is widely used in previous unsupervised methods is applied to warp the image in the source frame to the target frame, which is shown as the synthesized target image.
Because of the inconsistent visual information between the frames caused by motion and view-field changes, synthesized target images produced by view reconstruction are incomplete when compared with the real target image. These unreconstructed areas in the synthesized target images become a distinctive feature between the real and fake images for the discriminator, which will help the discriminator to distinguish them, thereby breaking the balance of the max-min game in adversarial learning. Moreover, because the visual field changes between the adjacent frames are inevitable, these distinctive features (unreconstructed regions) usually persist and cannot be eliminated by continuous training of the generator.
In this paper, to eliminate the effects of the unreconstructed regions on the discriminator, we present a Boolean mask to preprocess the real target images and construct the same unreconstructed regions on the real target image as the synthesized target image, as shown in Fig. \ref{fig:fig3}.
Therefore, the synthesized and real target images contain the same data distribution of these unreconstructed regions, which will have a positive effect on the training of the discriminator and thus the adversarial training process.

\subsection{Unsupervised depth and pose estimation}

\textbf{Generator:} Our generator consists of three networks for different tasks, as shown in Fig. \ref{fig:fig3}; a DepthNet for monocular depth prediction, a PoseNet for regressing the ego-motion between the target and source image, and a MaskNet for mask estimation. The view reconstruction algorithm is utilized to reconstruct the target image from a source image based on the output of the pose and depth networks. The mask predicted by MaskNet is utilized to eliminate the impacts of incomplete reconstruction regions on reconstruction loss and adversarial loss. For reconstruction loss, the effectiveness of mask has been demonstrated in recent works \cite{zhou2017unsupervised,li2019sequential}. For adversarial loss, we tackle this challenge by converting the mask into the Boolean mask to preprocess the target images and synthesize the same unreconstructed regions on target images as that on synthesized images.
We train the proposed architecture in an unsupervised manner, and the overall objective loss function is formulated as follows:
\begin{equation}
\mathcal{L}_{g} =  \alpha \mathcal{L}_{basic} + \varphi \mathcal{L}_{scale}+ \beta \mathcal{L}_{mask} + \gamma \mathcal{L}_{GAN},  \label{eq:1}
\end{equation}
where $\alpha$, $\varphi$, $\beta$ and $\gamma$ are the balance factors between the loss terms. $\mathcal{L}_{scale}$ refers to the proposed scale consistency loss, and $\mathcal{L}_{mask}$ is a regularization term to constrain the training of MaskNet, which is inspired by Zhou \textit{et al.} \cite{zhou2017unsupervised}. $\mathcal{L}_{GAN}$ denotes the adversarial loss. $\mathcal{L}_{basic}$ stands for the basic loss. The basic loss is used to assist the training process of the generator and consists of two parts, the traditional reconstruction loss $ \mathcal{L}_{rec}$ and the smoothness loss $\mathcal{L}_{smooth}$ :
\begin{equation}
\mathcal{L}_{basic} = \alpha_{1} \mathcal{L}_{rec} + \alpha_{2} \mathcal{L}_{smooth} ,  \label{eq:2}
\end{equation}
where $\alpha_{1}$ and $\alpha_{2}$ are the balance factors.

\textbf{Reconstruction Loss:} With the output of DepthNet and PoseNet, the target images can be reconstructed from the source images $I_{s}$ through the view reconstruction algorithm, which is widely used in previous unsupervised monocular depth estimation frameworks \cite{zhou2017unsupervised,yin2018geonet,bian2019depth}. The principle of this algorithm is based on the following projection function:
\begin{equation}
p_{s} \sim K \hat{T}_{t \to s} \hat{D}_{t}(p_{t})K^{-1}p_{t}, \label{eq:3}
\end{equation}
where $K$ stands for the camera intrinsics matrix. $\hat{D}_{t}$ denotes the depth estimation of the target image $I_{t}$, and $\hat{T}_{t \to s}$ represents the predicted 6-DOF transformation between the target image $I_{t}$ and the source image $I_{s}$. The pixels of two images $p_{t}, p_{s}$ establish the correspondence by a projection function. Then, we synthesize $\hat{I}_{s}$ by warping the image in source frame to target frame. Finally, the reconstruction loss is formulated as follows:
\begin{equation}
\mathcal{L}_{rec} = \alpha_{3} \dfrac{1-SSIM(I_{t},\hat{I}_{s})}{2} + (1-\alpha_{3}) ||I_{t}-\hat{I}_{s}||_{1}, \label{eq:4}
\end{equation}
where $\alpha_{3}$ refers to a balance factor, and SSIM \cite{wang2004image} is an index that shows the structural similarity between $I_{t}$ and $\hat{I}_{s}$.

\textbf{Smoothness Loss:} To filter out erroneous predictions and promote the representation of the geometric structure, a smoothness loss is designed to constrain the smoothness of the predicted depth map. Inspired by \cite{Yang2018lego}, we adopt the second-order differential for improving the smoothness:
\begin{equation}
\mathcal{L}_{smooth} = \sum_{p_{t}} |\nabla ^{2} D(p_{t})| (e^{-|\nabla ^{2} I({p}_{t})|})^{T}, \label{eq:5}
\end{equation}
where $\nabla ^{2}$ denotes the second-order differential operator, and $T$ is the transpose operation.

\begin{figure*}[t]
	\centering
	\includegraphics[width = 1.8\columnwidth]{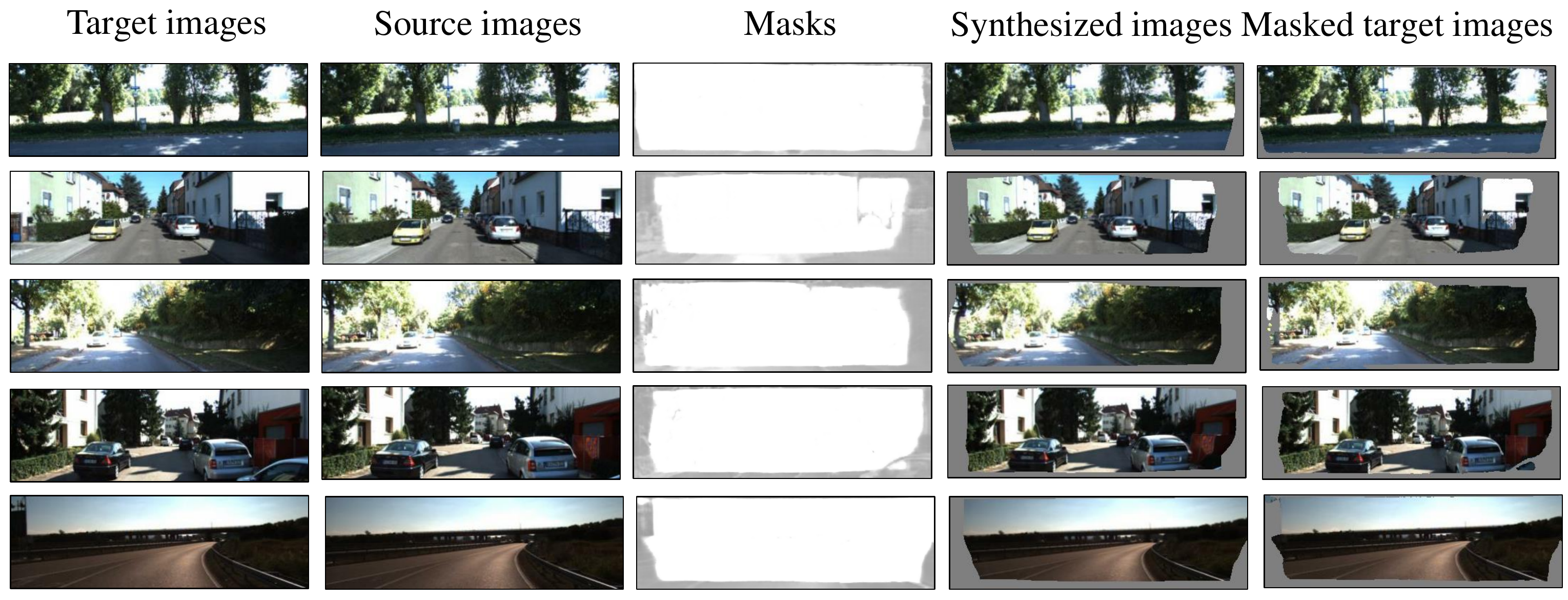}
	\caption{Qualitative results of MaskNet and Boolean mask processing. The ``masks'' are generated by MaskNet to predict unreconstructed regions on synthesized target images, and they are converted to the Boolean one to preprocess the target image. The ``synthesized images'' are reconstructed from source images by view reconstruction algorithm. The ``masked target images'' are preprocessed by Boolean mask and have the similar unreconstructed regions to synthesized images.}
	\label{fig:fig4}
\end{figure*}

\textbf{Traditional mask for reconstruction loss:} To eliminate the influence of the visual field changes and the dynamic objects on reconstruction loss, we use a MaskNet to predict these unreconstructed regions. During training, a MaskNet is designed to estimate the different regions between the real target image $I_{t}$ and the synthesized target image $\hat{I}_{s}$, formatted as $min[ M \times (I_{t}-\hat{I}_{s})]$, and a regularization term $\mathcal{L}_{mask}$ is used to constrain the MaskNet during training, which is similar to \cite{zhou2017unsupervised}. Our intention is to use the MaskNet to predict the regions that could not be reconstructed on synthesized images.
Considering the effect of incomplete reconstruction on the reconstruction loss, the mask $M$ predicted by MaskNet is introduced into $\mathcal{L}^{m}_{rec}$:
\begin{equation}
\mathcal{L}^{m}_{rec} = \sum_{p_{t}} M \mathcal{L}_{rec}. \label{eq:8}
\end{equation}
Our final basic model with the mask is formulated as follow:
\begin{equation}
\mathcal{L}^{m}_{basic} = \alpha_{1} \mathcal{L}^{m}_{rec} + \alpha_{2} \mathcal{L}_{smooth}.  \label{eq:9}
\end{equation}

\textbf{Scale Consistency Loss:} In unsupervised monocular frameworks, the depth and pose networks are trained by unlabelled short snippets. 
However, different snippets have different scale factors (compared with the ground truth) because there is no corresponding scale constraint loss applied. Therefore, previous pose and depth networks \cite{zhou2017unsupervised,yin2018geonet} cannot provide scale-consistent results between different snippets.
Bian \textit{et al.} \cite{bian2019depth} consider the scale-consistency of pose networks for generating full trajectories over long monocular videos. Similarly, this paper proposes a novel adaptive loss for better constraint on geometric and scale consistency between snippets, which is formulated as follows:
\begin{equation}
\mathcal{L}_{scale} = \alpha_{4} \dfrac{1-SSIM(D^{t}_{s},\hat{D}^{s}_{s})}{2} + (1-\alpha_{4}) ||D^{t}_{s}-\hat{D}^{s}_{s}||_{1},  \label{eq:6}
\end{equation}
\begin{equation}
D^{t}_{s}(p_{s}) \sim K \hat{T}_{t \to s} \hat{D}_{t}(p_{t})K^{-1}p_{t}, \label{eq:7}
\end{equation}
where $\alpha_{4}$ refers to a balance factor. $D^{t}_{s}(p_{s})$ is computed by the projection algorithm shown in Eq. (\ref{eq:7}), and it has the same scale information as $\hat{D}_{t}$. $\hat{D}_{t}$ stands for the predicted depth map of the target image. $\hat{D}_{s}$ is the predicted depth map of the source image (the target image of the next frame snippet). $\hat{D}^{s}_{s}$ is reconstructed from $\hat{D}_{s}$ by the warping algorithm, which is similar to the view reconstruction process in \cite{zhou2017unsupervised}, and it contains the same scale information as $\hat{D}_{s}$. Then, SSIM loss \cite{wang2004image} is adopted for the consistency of $D^{t}_{s}$ and $\hat{D}^{s}_{s}$ in such a way that the scale between the snippets is aligned.

\subsection{Boolean mask processing for the masked GAN}

\textbf{Original GAN:} consists of two components, a generator ($G$) and a discriminator ($D$) \cite{goodfellow2014generative}. The $D$ is designed to distinguish the real data $x$ from synthesized data $G(z)$ by learning the difference of data distribution. At the same time, the major task of the generator is to generate a set of data $G(z)$ with the same distribution as the real data to fool the discriminator. Therefore, a max-min game is played between the generator and discriminator, and the loss function of the original GAN \cite{goodfellow2014generative} is formulated as:
\begin{equation}
\begin{aligned}
\mathcal{L} &= \mathop{\emph{min}} \limits_{G}\mathop{\emph{max}} \limits_{D} V(D,G)\\
 &= \mathbb{E}_{x\sim p_{data}(x)}[\emph{log} D(x)]+\mathbb{E}_{z\sim p_{z}(z)}[\emph{log}(1- D(G(z)))],
\end{aligned}  \label{eq:10}
\end{equation}
where $ p_{data}(x)$ and $p_{z}(z)$ stand for the data distribution of $x$ and $z$, respectively. Finally, a mapping relationship between two data distributions is established through adversarial learning.

The unsupervised monocular depth estimation cannot feed the discriminator with a real depth map because there is no ground truth. To address this limitation, instead of distinguishing between the real and predicted depth map, the RGB images $\hat{I}_{s}$ synthesized by the view reconstruction are sent to discriminator together with the real target images $I_{t}$ in the monocular framework combined with GANs \cite{cs2018monocular,almalioglu2019ganvo}:
\begin{equation}
\begin{aligned}
&\mathcal{L}_{GAN} = \mathop{\emph{min}} \limits_{G}\mathop{\emph{max}} \limits_{D} V(D,G) \\
& = \mathbb{E}_{I_{t}\sim p(I_{t})}[\emph{log} D(I_{t})] + \mathbb{E}_{\hat{I}_{s}\sim p(\hat{I}_{s})}[\emph{log}(1- D(\hat{I}_{s}))],
\end{aligned} \label{eq:11}
\end{equation}
where $D(.)$ and $G(.)$ stand for the discriminator and generator of this paper.

Since the data distribution of the unreconstructed regions is unique to the synthetic images, which will affect the adversarial learning, we propose a Boolean mask to reduce the impact of the unreconstructed regions on the adversarial learning, and the same unreconstructed regions as synthesized images are produced on the target images.

\textbf{Boolean mask processing:} First, we transfer the floating-point mask $M$ predicted by MaskNet to a Boolean type $M_{b}$ by comparing it with a threshold $\theta$:
\begin{equation}
M_{b}(p) = \left\{ \begin{array}{ll}
0 & \textrm{if $|M(p)|\leq \theta$},\\
1 & \textrm{if $|M(p)| > \theta$},
\end{array} \right. \label{eq:12}
\end{equation}
where $p$ stands for the pixel index on the mask.
Then, to eliminate the distinctive feature (unreconstructed regions in synthesized images) between the synthesized and real target images, we use the Boolean mask $M_{b}$ to reweight the target image $I_{t}$ and get similar unreconstructed regions as the synthesized image $\hat{I}_{s}$ on the target image $I_{t}$:
\begin{equation}
I^{M_{b}}_{t} = M_{b}I_{t}. \label{eq:13}
\end{equation}
Besides, to prevent this Boolean mask processing step from introducing the new and particular noises into target images, which will also influence the training of discriminator, we do the same Boolean mask processing operation to the synthesized image $\hat{I}_{s}$:
\begin{equation}
\hat{I}^{M_{b}}_{s} = M_{b}\hat{I}_{s}. \label{eq:14}
\end{equation}

\textbf{Masked GAN:} After preprocessing by the Boolean mask, the masked target image $I^{M_{b}}_{t}$ (real) and synthesized target image $\hat{I}^{M_{b}}_{s}$ (fake) are sent to the discriminator, and based on \cite{goodfellow2014generative}, our final adversarial loss is formulated as follows:
\begin{equation}
\begin{aligned}
&\mathcal{L}^{M_{b}}_{GAN} = \mathop{\emph{min}} \limits_{G}\mathop{\emph{max}} \limits_{D} V(D,G) \\
 &= \mathbb{E}_{I^{M_{b}}_{t}\sim p(I^{M_{b}}_{t})}[\emph{log} D(I^{M_{b}}_{t})] + \mathbb{E}_{\hat{I}^{M_{b}}_{s}\sim p(\hat{I}^{M_{b}}_{s})}[\emph{log}(1- D(G(I_{t},I_{s})))] \\
& = \mathbb{E}_{I^{M_{b}}_{t}\sim p(I^{M_{b}}_{t})}[\emph{log} D(I^{M_{b}}_{t})] + \mathbb{E}_{\hat{I}^{M_{b}}_{s}\sim p(\hat{I}^{M_{b}}_{s})}[\emph{log}(1- D(\hat{I}^{M_{b}}_{s}))],
\end{aligned} \label{eq:15}
\end{equation}
where $I^{M_{b}}_{t}$ and $\hat{I}^{M_{b}}_{s}$ are calculated by Boolean mask processing, Eq.\ref{eq:13} and Eq. \ref{eq:14}.

\textbf{Floating-point mask processing:}
To verify the effectiveness of Boolean mask processing, instead of transferring the mask to the Boolean one, the floating-point mask $M_{f}=M$ predicted by MaskNet is directly used to preprocess the synthesized and real target images ($\hat{I}_{s}$, $I_{t}$ ):
\begin{equation}
\hat{I}^{M_{f}}_{s} = M_{f}\hat{I}_{s},\qquad I^{M_{f}}_{t} = M_{f}I_{t}. \label{eq:16}
\end{equation}
Therefore, based on \cite{goodfellow2014generative}, the adversarial loss combined with floating-point mask processing is formulated as:
\begin{equation}
\begin{aligned}
&\mathcal{L}^{M_{f}}_{GAN} = \mathop{\emph{min}} \limits_{G}\mathop{\emph{max}} \limits_{D} V(D,G) \\
& = \mathbb{E}_{I^{M_{f}}_{t}\sim p(I^{M_{f}}_{t})}[\emph{log} D(I^{M_{f}}_{t})] + \mathbb{E}_{\hat{I}^{M_{f}}_{s}\sim p(\hat{I}^{M_{f}}_{s})}[\emph{log}(1- D(\hat{I}^{M_{f}}_{s}))],
\end{aligned} \label{eq:17}
\end{equation}
where $I^{M_{f}}_{t}$ and $\hat{I}^{M_{f}}_{s}$ are calculated by floating-point mask processing, shown in Eq.(\ref{eq:16}).



\begin{table*}[]
	
	\scriptsize
	
	\centering
	
	\caption{ \textit{Ablation study results on the KITTI raw dataset \cite{geiger2013vision}. Based on the basic loss function ($\mathcal{L}_{basic}$, Eq. (\ref{eq:2})), scale consistency loss ($\mathcal{L}_{scale}$, Eq. (\ref{eq:6})), adversarial learning ($\mathcal{L}_{GAN}$, Eq. (\ref{eq:11})), masks for reconstruction loss ($\mathcal{L}^{m}_{basic}$, Eq. (\ref{eq:9}), $\mathcal{L}_{mask}$), floating-point mask processing for adversarial learning ($\mathcal{L}^{M_{f}}_{GAN}$, Eq. (\ref{eq:17})), and Boolean mask processing for adversarial learning ($\mathcal{L}^{M_{b}}_{GAN}$, Eq. (\ref{eq:15})) are introduced sequentially into training.} }
	
	\label{Tab01}
	\renewcommand\arraystretch{1.2}
	\begin{tabular}{c|c|cccc|ccc}
		\toprule
		\multicolumn{2}{c}{}& \multicolumn{4}{c}{Lower is better} & \multicolumn{3}{c}{Higher is better}  \\
		\cmidrule(r){3-6} \cmidrule(r){7-9}
		\hline
         \multicolumn{9}{c}{Ablation study based on different loss function}\\
		\hline
		Method  &Resolution	&  Abs Rel      &  Sq Rel    &   RMSE 	&  RMSE log     &   $\delta < 1.25^{1}$		&  $\delta < 1.25^{2}$      &  $\delta < 1.25^{3}$ \\
		\hline
        $\mathcal{L}_{\textrm{basic}}$  & 416$\times$128 &  0.154  &  1.287 &   5.918	&  0.236  &   0.796	&  0.925    & 0.970 \\
		$\mathcal{L}_{\textrm{basic}}$ + $\mathcal{L}_{\textrm{scale}}$ & 416$\times$128 &  0.152  &  1.166  &   5.552	&  0.225   &   0.794	&  0.934    & 0.974 \\
		$\mathcal{L}_{\textrm{basic}}$ + $\mathcal{L}_{\textrm{scale}}$ + $\mathcal{L}_{\textrm{GAN}}$   & 416$\times$128 &  0.148  &  1.104   &   5.667	&  0.227   &   0.801	&  0.932    & 0.974 \\
		$\mathcal{L}^{m}_{\textrm{basic}}$ + $\mathcal{L}_{\textrm{scale}}$ + $\mathcal{L}_{\textrm{GAN}}$ + $\mathcal{L}_{\textrm{mask}}$	& 416$\times$128 & \textbf{0.145}  &  1.095   &   5.601	&  0.223 &   0.803	&  0.932    & 0.975 \\
		$\mathcal{L}^{m}_{\textrm{basic}}$ + $\mathcal{L}_{\textrm{scale}}$ + $\mathcal{L}^{M_{f}}_{\textrm{GAN}}$ + $\mathcal{L}_{\textrm{mask}}$	& 416$\times$128
&  0.147  &  1.086  &   5.555 	&  0.223  &   0.799	&  0.934    & 0.976 \\
		$\mathcal{L}^{m}_{\textrm{basic}}$ + $\mathcal{L}_{\textrm{scale}}$ + $\mathcal{L}^{M_{b}}_{\textrm{GAN}}$ + $\mathcal{L}_{\textrm{mask}}$ 	& 416$\times$128  &   0.146 &   \textbf{1.084}   &    \textbf{5.445}	&   \textbf{0.221}   &    \textbf{0.807}	&   \textbf{0.936}   &  \textbf{0.976} \\
		\hline
         \multicolumn{9}{c}{Comparison between different mask networks}\\
		\hline
		Mask network  &Resolution	&  Abs Rel      &  Sq Rel    &   RMSE 	&  RMSE log     &   $\delta < 1.25^{1}$		&  $\delta < 1.25^{2}$      &  $\delta < 1.25^{3}$ \\
		\hline
        MaskNet (Ours) & 416$\times$128 &   \textbf{0.146} &   \textbf{1.084}   &    \textbf{5.445}	&   \textbf{0.221}   &    \textbf{0.807}	&   \textbf{0.936}   &  \textbf{0.976} \\
		Mask network ([2])  & 416$\times$128 &  0.149  &  1.121 &   5.580	&  0.225  &   0.799	&  0.932    & 0.974 \\
		\hline
         \multicolumn{9}{c}{Parameter optimization}\\
		\hline
		Weight  &Resolution	&  Abs Rel      &  Sq Rel    &   RMSE 	&  RMSE log     &   $\delta < 1.25^{1}$		&  $\delta < 1.25^{2}$      &  $\delta < 1.25^{3}$ \\
		\hline
        $\gamma=0.01$  & 416$\times$128 &  0.150  &  1.107 &   5.594	&  0.227  &   0.799	&  0.931    & 0.974 \\
		$\gamma=0.001$  & 416$\times$128 &  0.148  &  1.091  &   5.536	&  0.223   &   0.802	&  0.934    & 0.976 \\
		$\gamma=0.0001$  & 416$\times$128 &  0.146  &  1.081   &   5.544	&  0.222   &   0.805	&  0.934    & 0.975 \\
		$\gamma=0.00001$  & 416$\times$128 &  0.147  &  1.103   &   5.520 	&  0.224  &   0.805&  0.934    & 0.975 \\
        \hline
		$\beta=0.4$	  & 416$\times$128 &  \textbf{0.146}  &  \textbf{1.081}   &   5.544	&  0.222   &   0.805	&  0.934    & 0.975 \\
		$\beta=0.3$   & 416$\times$128  &   \textbf{0.146} &   1.084   &    \textbf{5.445}	&   \textbf{0.221}   &    \textbf{0.807}	&   \textbf{0.936}   &  \textbf{0.976} \\
		$\beta=0.25$	  & 416$\times$128 &  0.147  &  1.091   &   5.448	&  0.222   &   0.804	&  0.936    & 0.976 \\
		\bottomrule
		
	\end{tabular}
	
\end{table*}

\section{Experiments}

\subsection{Implementation Details}
\textbf{Network architecture: } Our unsupervised method mainly consists of two modules, the generator and the discriminator, and the model can be divided into three subnetworks, the MaskNet coupled with the PoseNet, the DepthNet, and the discriminator network. For the DepthNet, we follow previous work \cite{ranjan2019competitive,bian2019depth} and take the DispResNet \cite{yin2018geonet} as our DepthNet. The DepthNet takes a single image as input and outputs a depth map in an end-to-end manner.
For the PoseNet and MaskNet, we directly adopt the coupled framework proposed in \cite{zhou2017unsupervised,li2019sequential}, while the main difference is that we add the skip-connections between the corresponding encoding and decoding layers to improve the performance of the mask prediction. This coupled framework takes a concatenated RGB image snippet as input, which consists of one target image and two or four source images. The PoseNet predicts the 6 DOF poses between the target and source images, and the MaskNet predicts the masks to handle the occlusions and visual field changes between the target and source images.
The architecture of discriminator network is designed with reference to the encoder of DispNet \cite{mayer2016large}. The real target images and synthesized target images are both preprocessed by a Boolean mask processing step before being sent to the discriminator. During training, the discriminator outputs the probability that the input image is real or fake, and the adversarial learning between the generator and discriminator improves the training of the depth and pose networks.

\textbf{Training detail:}
The proposed adversarial learning framework is implemented using TensorFlow \cite{abadi2016tensorflow}. For the depth prediction task, we train and test our DepthNet on the KITTI raw dataset \cite{geiger2013vision} by using Eigen's split \cite{eigen2014depth}. For the pose prediction task, following \cite{zhan2018unsupervised}, we train and test our PoseNet on the KITTI odometry dataset \cite{geiger2013vision}, where the sequences 00-08 are used for training and sequences 09-10 are used for testing. The setup of our training and testing sets is the same as the ones in previous related studies \cite{zhou2017unsupervised,yin2018geonet,bian2019depth,zhan2018unsupervised}.
We regard a snippet of three (for training PoseNet) or five (for training DepthNet) sequential video frames as a training sample, which consists of one target image (middle frame) and two or four source images. We have done experiments with two different resolutions, $416 \times 128$ and $640 \times 192$. We train our framework in 50 epochs and obtain 10 randomly sampled batches in one epoch for validation, which is different from previous work in which the framework is trained in 100 epochs \cite{xiong2020selfsupervised} or 200 epochs \cite{bian2019depth} and 1000 randomly sampled batches are obtained in one epoch for validation, and thus, it indicates that our model shows a good convergence rate. Following \cite{bian2019depth,xiong2020selfsupervised}, we also pre-train our networks on Cityscapes \cite{Cordts2016Cityscapes} and fine-tune it on KITTI \cite{geiger2013vision}, each for 50 epochs.

The training of our networks is based on a single RTX 8000 GPU. During training, the image resolution is resized to $416\times128$ or $640\times192$, the weights of the generator and discriminator are optimized by ADAM optimizers \cite{kingma2014adam} with $\beta_{1}=0.9$, $\beta_{2}=0.999$, and the learning rate is 0.0002. During training, we adopt $\alpha_{1}=1.0$, $\alpha_{2}=0.5$, $\alpha_{3}=0.85$, $\alpha_{4}=0.2$, $\varphi=0.5$, $\beta=0.3$, $\gamma=0.0001$, $\theta=0.9$. Our parameters are set based on the experiments or related work \cite{yin2018geonet}, and we show some experimental results of hyper-parameter optimization in Table \ref{Tab01}.

\begin{table*}[]
	
	\scriptsize
	
	\centering
	
	\caption{ \textit{Comparison with the  methods using GANs for unsupervised depth and VO estimation. ``\textbf{SC}" stands for whether the scale-consistent is considered in their work. $^{\textcolor{red}{*}}$ - denotes that their depth map is not generated from a single image in an end-to-end manner. }}
	
	\label{Tab02}
	
	\begin{tabular}{c|c|c|cccc|ccc}
		
		\toprule
		\multicolumn{3}{c}{}& \multicolumn{4}{c}{Lower is better} & \multicolumn{3}{c}{Higher is better}  \\
		\cmidrule(r){1-3}\cmidrule(r){4-7} \cmidrule(r){8-10}
		\hline
		Method   	&  Resolution     &   \textbf{SC}		&  Abs Rel      &  Sq Rel    &   RMSE 	&  RMSE log     &   $\delta < 1.25^{1}$		&  $\delta < 1.25^{2}$      &  $\delta < 1.25^{3}$ \\
		\hline
		Kumar \textit{et al.} \cite{cs2018monocular} 	&  384$\times$128     &   $\times$			&  0.211  &  1.980   &   6.154	&  0.264   &   0.732	&  0.898    & 0.959 \\
		GANVO $^{\textcolor{red}{*}}$ \cite{almalioglu2019ganvo}  &  416$\times$128    &   $\times$			&  0.150 &  1.141  &   5.448	&  \textbf{0.216}   &   0.808	& \textbf{0.939}    & 0.975 \\
        Li \textit{et al.} $^{\textcolor{red}{*}}$ \cite{li2019sequential}  &  416$\times$128     &   $\surd$		&  0.150 &  1.127  &   5.564	&  0.229   &   \textbf{0.823}	& 0.936    & 0.974 \\
		Ours   	&  416$\times$128     &   $\surd$		&  \textbf{0.146} &  \textbf{1.084}   &  \textbf{5.445}	&  0.221   &  0.807	& 0.936   & \textbf{0.976} \\
		\hline
		Ours  &  640$\times$192     &   $\surd$	& \textbf{0.139} &  \textbf{1.034}  &   \textbf{5.264}	& \textbf{0.214}   &   0.821	&   \textbf{0.942}    & \textbf{0.978} \\
		
		\bottomrule
		
	\end{tabular}
\end{table*}

\textbf{Evaluation metrics:}
For depth evaluation, the commonly used evaluation metrics proposed by Eigen \textit{et al.} \cite{eigen2014depth} are used in this paper to compare our method with others fairly; these include five evaluation indicators: \textbf{ RMSE, RMSE log, Abs Rel, Sq Rel, Accuracy}:

\begin{itemize}
	\item $\textbf{RMSE} = \sqrt{\frac{1}{|N|}\sum_{i\in N}\parallel d_{i}-d_{i}^{*} \parallel^{2}}$,
	
	\item $\textbf{RMSE log} = \sqrt{\frac{1}{|N|}\sum_{i\in N}\parallel \log (d_{i})- \log (d_{i}^{*}) \parallel^{2}}$,
	
	\item $\textbf{Abs Rel} = \frac{1}{|N|}\sum_{i\in N}\frac{\mid d_{i}-d_{i}^{*}\mid}{d_{i}^{*}}$,\quad
	\item $\textbf{Sq Rel} = \frac{1}{|N|}\sum_{i\in N}\frac{\parallel d_{i}-d_{i}^{*}\parallel^{2}}{d_{i}^{*}}$,
	
	\item \textrm{\textbf{Accuracy:} $\%$ of $d_{i}$ s.t. } $\max(\frac{d_{i}}{d_{i}^{*}}, \frac{d_{i}^{*}}{d_{i}}) = \delta < thr$,
\end{itemize}
where $d_{i}$ and $d_{i}^{*}$ denote the predicted depth of pixel $i$ and the corresponding ground truth, and $N$ denotes the total number of pixels that correspond to the ground truth depth value. $thr$ denotes a threshold, which is always set to $1.25^{1}$, $1.25^{2}$, and $1.25^{3}$.

For the trajectory evaluation, the generated full trajectory is evaluated by the standard evaluation metrics provided in the dataset \cite{geiger2013vision}, including a translation error metric $t_{err}(\%)$ and a rotation error metric $r_{err}(^{\circ}/ 100 m)$. Compared with the 5-frame pose evaluation that proposed in \cite{zhou2017unsupervised}, the evaluation metrics in this paper are more widely used in traditional VO methods and are more meaningful.

\begin{figure}[t]
	\centering
	\includegraphics[width = \columnwidth]{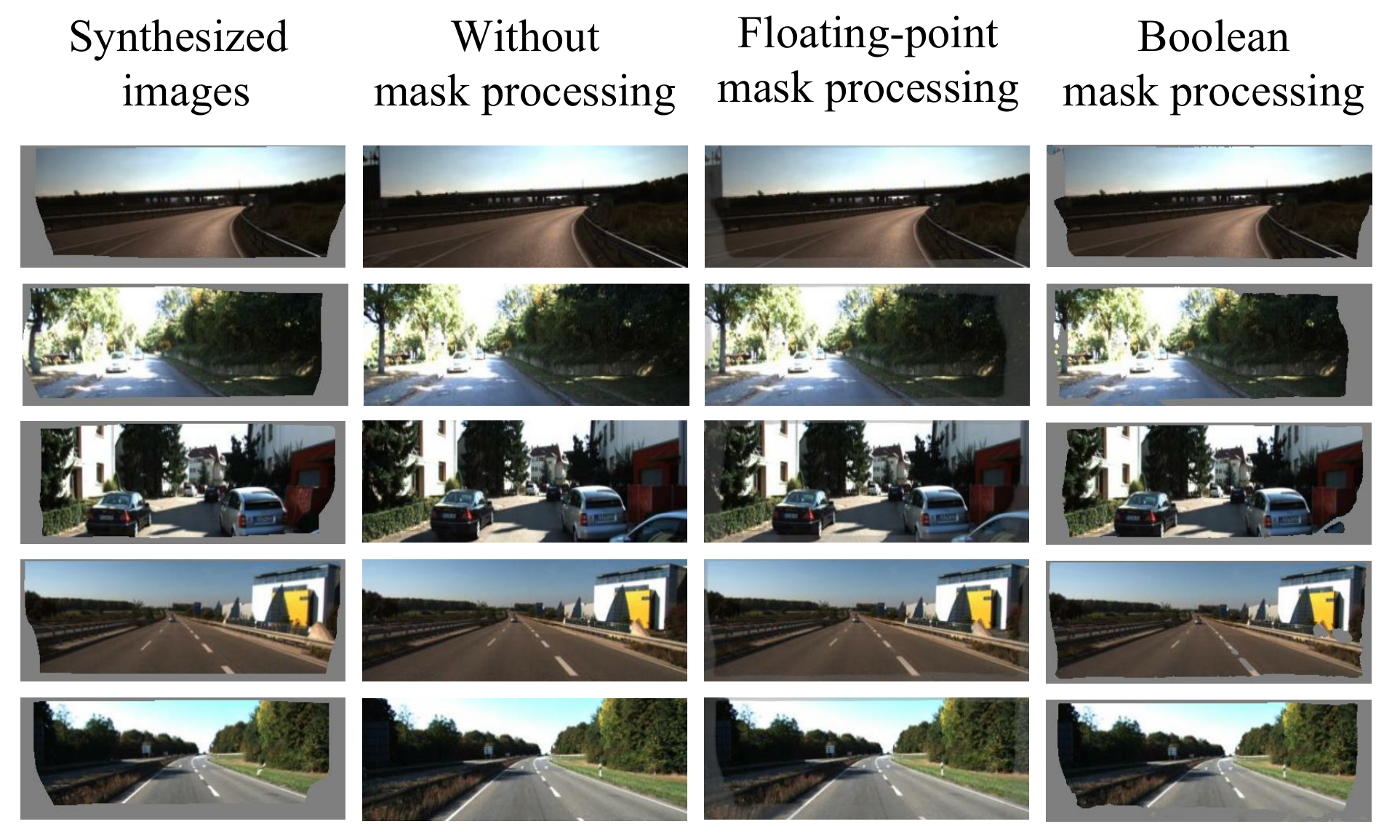}
	\caption{Visual results of different kinds of mask processing. In each row, images from left-to-right show the synthesized targte image, the real target image, the target image preprocessed by floating-point mask, and the target image preprocessed by Boolean mask. Floating-point mask processing can only dilute the view while Boolean mask processing is able to construct empty regions that are almost identical to synthesized images.}
	\label{fig:fig5}
\end{figure}

\subsection{Monocular Depth Estimation}

\textbf{Ablation study:} In this section, we first conduct a series of ablation experiments to validate the efficacy of our proposed framework, as shown in Table \ref{Tab01}. The ``$\mathcal{L}_{\textrm{basic}}$'' denotes that our framework is trained by the basic loss $\mathcal{L}_{\textrm{basic}}$ (Eq. (\ref{eq:2})), which consists of the reconstruction loss ($\mathcal{L}_{\textrm{rec}}$, Eq. (\ref{eq:4})) and smoothness loss ($\mathcal{L}_{\textrm{smooth}}$, Eq. (\ref{eq:5})). Then, scale consistency loss ($\mathcal{L}_{\textrm{scale}}$, Eq. (\ref{eq:6})), adversarial learning ($\mathcal{L}_{\textrm{GAN}}$, Eq. (\ref{eq:11})), masks for reconstruction loss ($\mathcal{L}^{m}_{\textrm{basic}}$, Eq. (\ref{eq:9}), $\mathcal{L}_{mask}$), and Boolean mask processing ($\mathcal{L}^{M_{b}}_{\textrm{GAN}}$, Eq. (\ref{eq:15})) are introduced sequentially into training. Although both the mask and Boolean mask processing (BMP) in Table \ref{Tab01} are based on the output of MaskNet, the mask is to reduce the effect of the unreconstructed regions on the reconstruction loss, while BMP is applied to eliminate the impact of the unreconstructed regions on the adversarial loss.

\begin{table*}[]
	
	\scriptsize
	
	\centering
	
	\caption{ \textit{Monocular depth results on KITTI \cite{geiger2013vision} by the split of Eigen \textit{et al.} \cite{eigen2014depth}. For training, Mono. /stereo denotes that networks trained by monocular sequences / stereo image pairs. ``-ResNet" stands for the methods with DispResNet \cite{yin2018geonet}. We show the best results in bold.}}
	
	\label{Tab03}
	
	\begin{tabular}{c|c|c|c|c|cccc|ccc}
		
		\toprule
		\multicolumn{5}{c}{}& \multicolumn{4}{c}{Lower is better} & \multicolumn{3}{c}{Higher is better}  \\
		\cmidrule(r){1-5}\cmidrule(r){6-9} \cmidrule(r){10-12}
		\hline
		Method   &  Dataset & Supervision 	&  Resolution     &   Cap		&  Abs Rel      &  Sq Rel    &   RMSE 	&  RMSE log     &   $\delta < 1.25^{1}$		&  $\delta < 1.25^{2}$      &  $\delta < 1.25^{3}$ \\
		\hline
		SfMLearner \cite{zhou2017unsupervised} & K  &   Mono. 	&  416$\times$128   &   80m		&  0.208  &  1.768   &   6.865 	&  0.283   &   0.678	&  0.885    & 0.957 \\
		Yang \textit{et al.} \cite{yang2017unsupervised} & K &   Mono. 	&  416$\times$128     &   80m		&  0.182  &  1.481   &   6.501 	&  0.267   &   0.725	&  0.906    & 0.963 \\
		Vid2depth \cite{mahjourian2018unsupervised} & K &   Mono. 	& 416$\times$128    &   80m		&  0.163  &  1.240   &   6.220 	&  0.250   &   0.762	&  0.916    & 0.968 \\
		GeoNet-ResNet. \cite{yin2018geonet} & K &   Mono. 	&  416$\times$128    &   80m		&  0.155  &  1.296   &   5.857 	&  0.233   &   0.793	&  0.931    & 0.973 \\
		Wang \textit{et al.} \cite{wang2019unsupervised} & K &   Mono. 	&  416$\times$128     &   80m		&  0.154  &  1.163   &   5.700 	&  0.229   &   0.792	&  0.932    & 0.974 \\
		SC-SfM-ResNet  \cite{bian2019depth}& K &   Mono. 	&  416$\times$128     &   80m		&  0.149 &  1.137   &   5.771	&  0.230   &   0.799	&  0.932    & 0.973 \\
		Xiong \textit{et al.}  \cite{xiong2020selfsupervised}& K &   Mono. 	&  416$\times$128     &   80m		&  0.148 &  \textbf{1.077}   &   5.506	&  0.228   &   0.806	&  0.934    & 0.973 \\
		Ours  & K  &   Mono 	&  416$\times$128     &   80m		&  \textbf{0.146} &  1.084   &   \textbf{5.445}	&  \textbf{0.221}   &  \textbf{0.807}	&  \textbf{0.936}   & \textbf{0.976} \\
		\hline
		DF-Net \cite{zou2018df} & K&   Mono. 	&  576$\times$160     &   80m		&  0.150  &  1.124   &   5.507 	&  0.223   &   0.806	&  0.933    & 0.973 \\
		CC  \cite{ranjan2019competitive}& K &   Mono. 	&  832$\times$256     &   80m		&  0.140 &  1.070   &  5.326	&  0.217   &   0.826	&  0.941    & 0.975 \\
		SC-SfM  \cite{bian2019depth} & K&   Mono. 	&  832$\times$256     &   80m		& \textbf{0.137}  &  1.089   &   5.439	& 0.217   &   \textbf{0.830}	&  \textbf{0.942}    & 0.975 \\
		Xiong \textit{et al.}  \cite{xiong2020selfsupervised} & K&   Mono. 	&  832$\times$256     &   80m		& 0.140 &  1.061  &   5.309	& 0.219   &   0.823	&  0.940    & 0.976 \\
		Ours & K &   Mono. 	&  640$\times$192     &   80m		& 0.139 &  \textbf{1.034}  &   \textbf{5.264}	& \textbf{0.214}   &   0.821	&  \textbf{0.942}    & \textbf{0.978} \\
		\hline
        \hline
		DF-Net \cite{zou2018df} & CS+K&   Mono. 	&  576$\times$160     &   80m		&  0.146  &  1.182  &   5.215 	&  0.213   &   0.818	&  0.943    & 0.978 \\
		CC  \cite{ranjan2019competitive}& CS+K &   Mono. 	&  832$\times$256     &   80m		&  0.139 &  1.032   &  5.199	&  0.213   &   0.827	&  0.943    & 0.977 \\
		SC-SfM  \cite{bian2019depth} & CS+K&   Mono. 	&  832$\times$256     &   80m		& 0.128  &  1.047   &   5.234	& 0.208  &   0.846	&  0.947    & 0.976 \\
		Xiong \textit{et al.}  \cite{xiong2020selfsupervised} & CS+K&   Mono. 	&  832$\times$256     &   80m		& \textbf{0.126} &  \textbf{0.902}  &   \textbf{5.052}	& \textbf{0.205}   &   \textbf{0.851}	&  \textbf{0.950}    & \textbf{0.979} \\
		Ours & CS+K &   Mono. 	&  640$\times$192     &   80m		& 0.135 &  1.026  &   5.153	& 0.210   &   0.833	&  0.945    & \textbf{0.979} \\
		\bottomrule
		
	\end{tabular}
\end{table*}

From the results shown in Table \ref{Tab01}, comparing lines 1 and 2 in the table, the proposed scale consistency loss $\mathcal{L}_{\textrm{scale}}$ is conducive to improvement in the depth estimation.
Comparing lines 2 and 3, the introduction of adversarial learning (GAN) does not effectively improve the accuracy of the depth network, $i.e.$, the introduced adversarial learning does not play a key role in the training of DepthNet.
In addition, we achieve a better result when the proposed $\mathcal{L}^{M_{b}}_{\textrm{GAN}}$ is used to reduce the impact of the unreconstructed regions on adversarial learning.
Therefore, we believe that the unreconstructed regions influence the performance of adversarial learning.
Then, the mask predicted by MaskNet is introduced into the framework, which is shown as ``$\mathcal{L}^{m}_{\textrm{basic}}$ + $\mathcal{L}_{\textrm{scale}}$ + $\mathcal{L}_{\textrm{GAN}}$ + $\mathcal{L}_{\textrm{mask}}$''. From the evaluation metrics, the depth estimation accuracy of ``$\mathcal{L}^{m}_{\textrm{basic}}$ + $\mathcal{L}_{\textrm{scale}}$ + $\mathcal{L}_{\textrm{GAN}}$ + $\mathcal{L}_{\textrm{mask}}$'' outperforms that of ``$\mathcal{L}_{\textrm{basic}}$ + $\mathcal{L}_{\textrm{scale}}$ + $\mathcal{L}_{\textrm{GAN}}$'', which means that the adopted mask improves the training process of the network.
With the qualitative results on the mask in Fig. \ref{fig:fig4}, the unreconstructed regions of the synthesized images are accurately predicted in the masks. Therefore, the reason for accuracy improvement in the depth estimation is that the mask can significantly mitigate the influence of the unreconstructed regions on the reconstruction loss.
Afterwards, we adopt the BMP in this paper to mask the real target images (real) and synthesized target images (fake) before sending them to the discriminator, which is shown as ``$\mathcal{L}^{m}_{\textrm{basic}}$ + $\mathcal{L}_{\textrm{scale}}$ + $\mathcal{L}^{M_{b}}_{\textrm{GAN}}$ + $\mathcal{L}_{\textrm{mask}}$'', and the best result among the four cases is achieved.
In addition, the qualitative results of the BMP shown in Fig. \ref{fig:fig4} also demonstrate that our proposed BMP can construct similar unreconstructed regions in target images, which helps to reduce the impact of the unreconstructed regions on the adversarial learning and plays a key role in adversarial learning.
In summary, this ablation study indicates the effectiveness of our proposed modules.

\textbf{Comparison of different mask networks:} Since the mask predicted by the MaskNet is very important in this paper and plays a key role in the adversarial loss and reconstructing loss, we applied some modifications (skip connection between encoder and decoder) on the mask network proposed in \cite{zhou2017unsupervised} to improve its performance. As shown in Table \ref{Tab01}, the experimental result shows that these modifications effectively improve the performance of our framework, which also illustrates the importance of the MaskNet to our framework.

\textbf{Comparison of different mask processing:}
To further verify the ability of the Boolean mask, we compare the GAN processed by the BMP ($\mathcal{L}^{M_{b}}_{\textrm{GAN}}$) with cases of GANs processed by floating-point mask (FMP) ($\mathcal{L}^{M_{f}}_{\textrm{GAN}}$) and the original GAN without mask processing ($\mathcal{L}_{GAN}$); the results are shown in lines 4, 5 and 6 of Table \ref{Tab01}.
Although the FMP cannot create similar unreconstructed regions on real target images as synthesized target images, it can be seen from Table \ref{Tab01} that FMP plays a positive role in improving the performance of adversarial learning when compared with the original GAN.
According to the examples of the different mask processing results shown in Fig. \ref{fig:fig5}, the effect of the FMP is not as obvious as that of the BMP.
In addition, the experimental results in Table \ref{Tab01} show that both the FMP and BMP are effective in accuracy improvement of the depth estimation, and the BMP works the best.

\begin{figure}[t]
	\centering
	\includegraphics[width = \columnwidth]{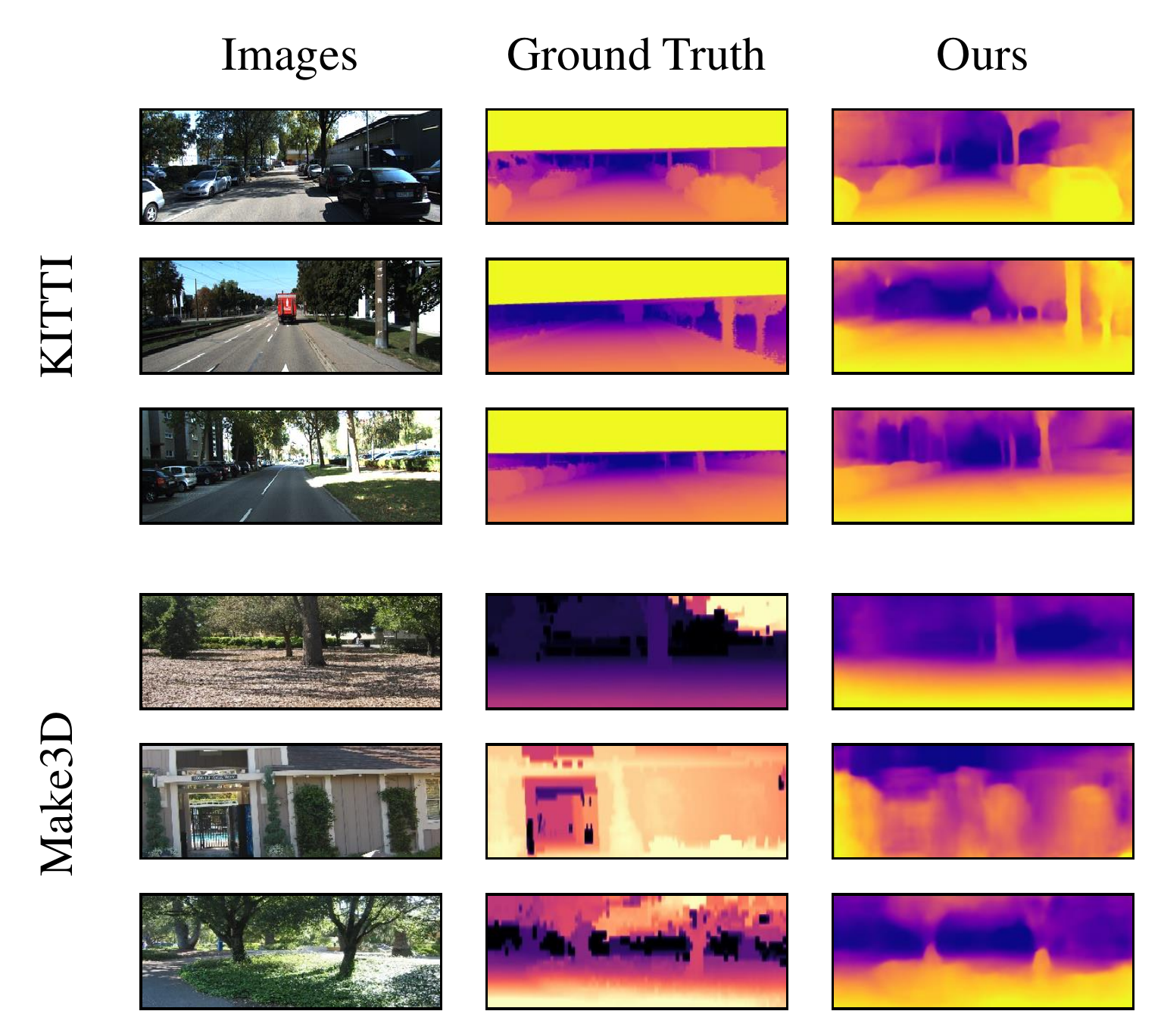}
	\caption{Qualitative results of our depth networks, which is trained in an unsupervised manner with a resolution of $416\times128$. The geometric structures in the scenes, such as boles and cars, can be effectively presented by our depth model from single images.}
	\label{fig:fig6}
\end{figure}

\begin{figure*}[htbp]
	\centering
	\subfigure[Trajectory of Seq. 09 with scale correction ]{
		\includegraphics[width = 0.47\columnwidth]{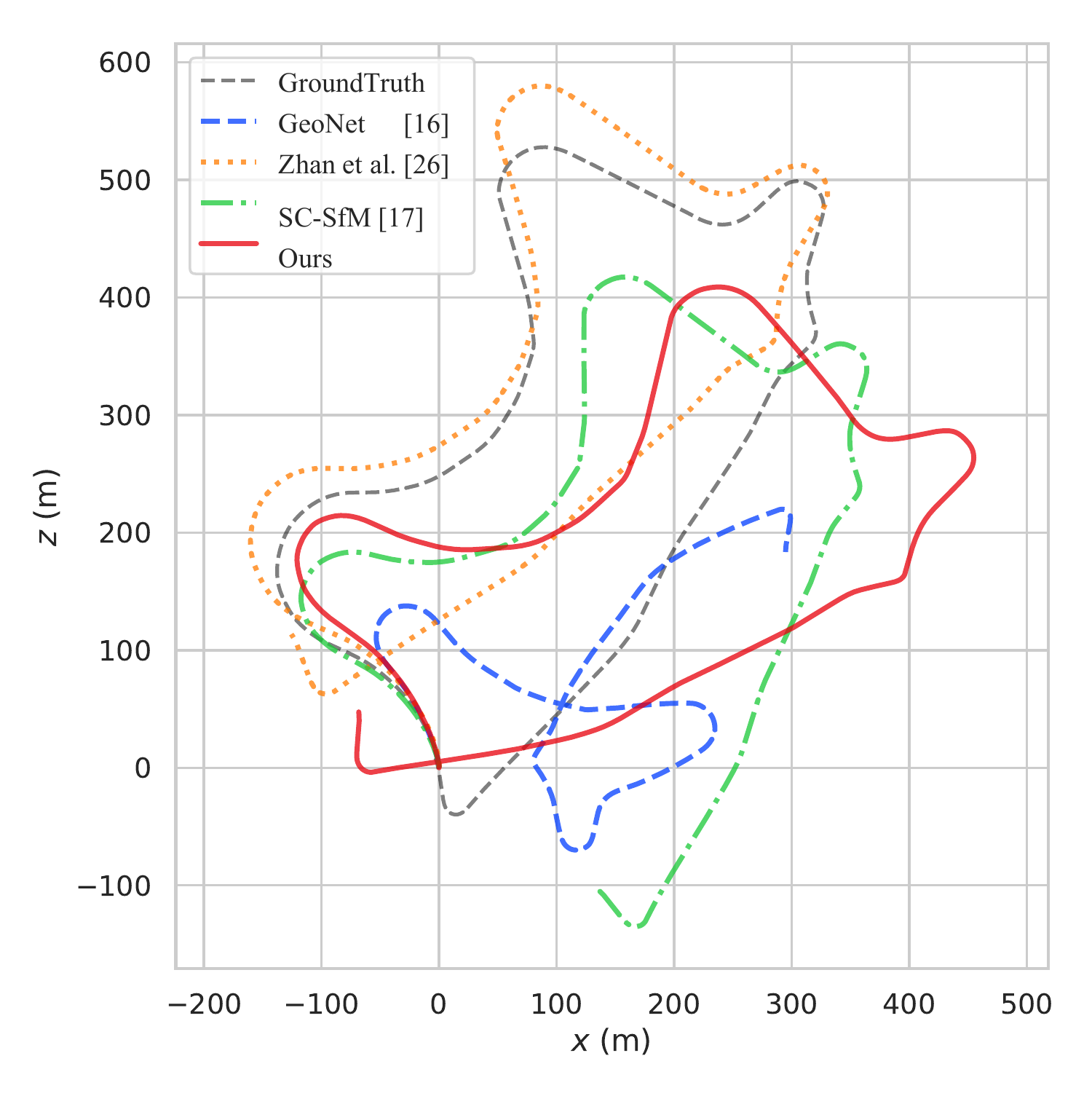}
	}
	\subfigure[Trajectory of Seq. 10 with scale correction]{
		\includegraphics[width = 0.47\columnwidth]{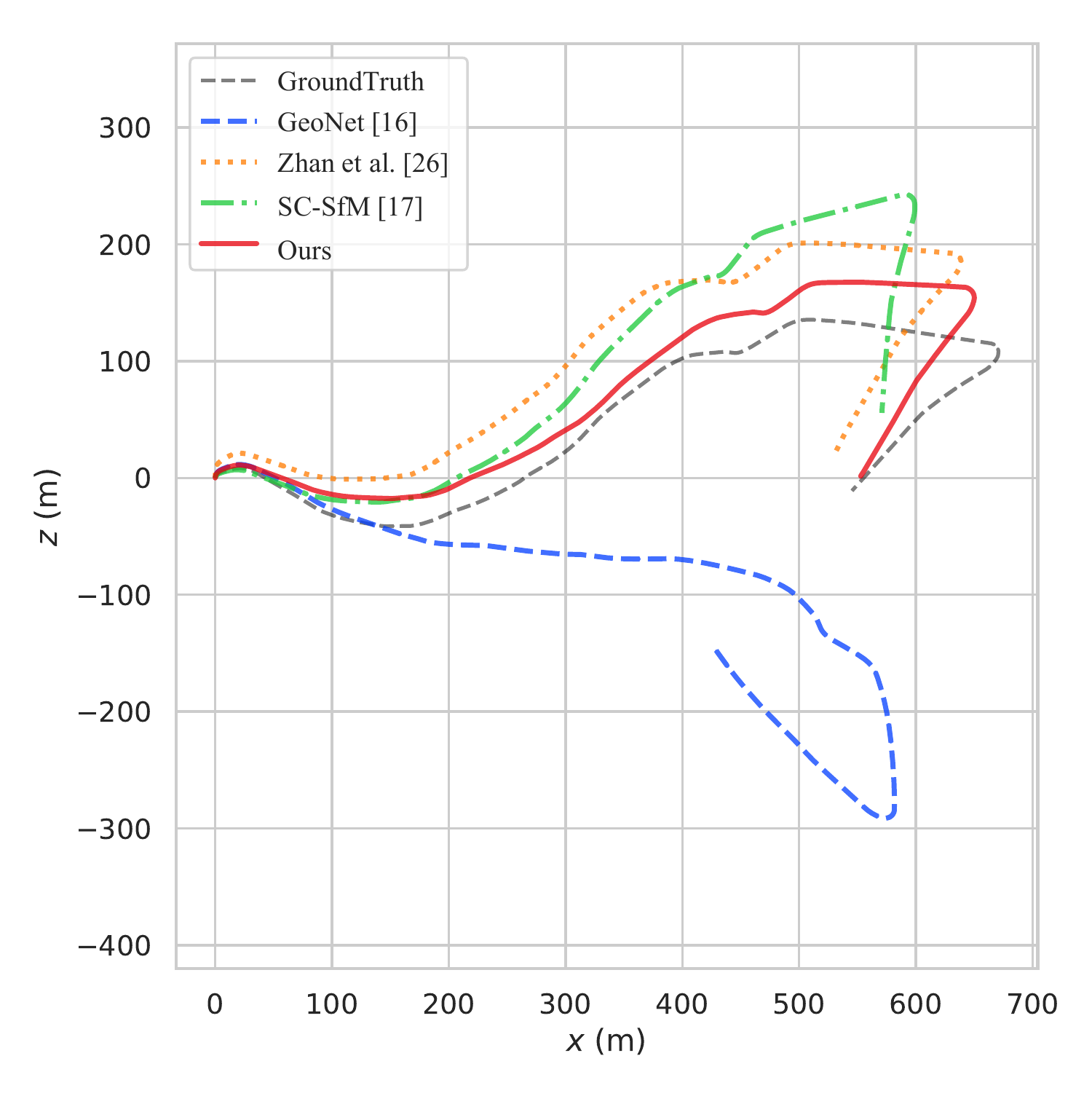}
	}
	\subfigure[Trajectory of Seq. 09 with SE(3) Umeyama alignment]{
		\includegraphics[width = 0.47\columnwidth]{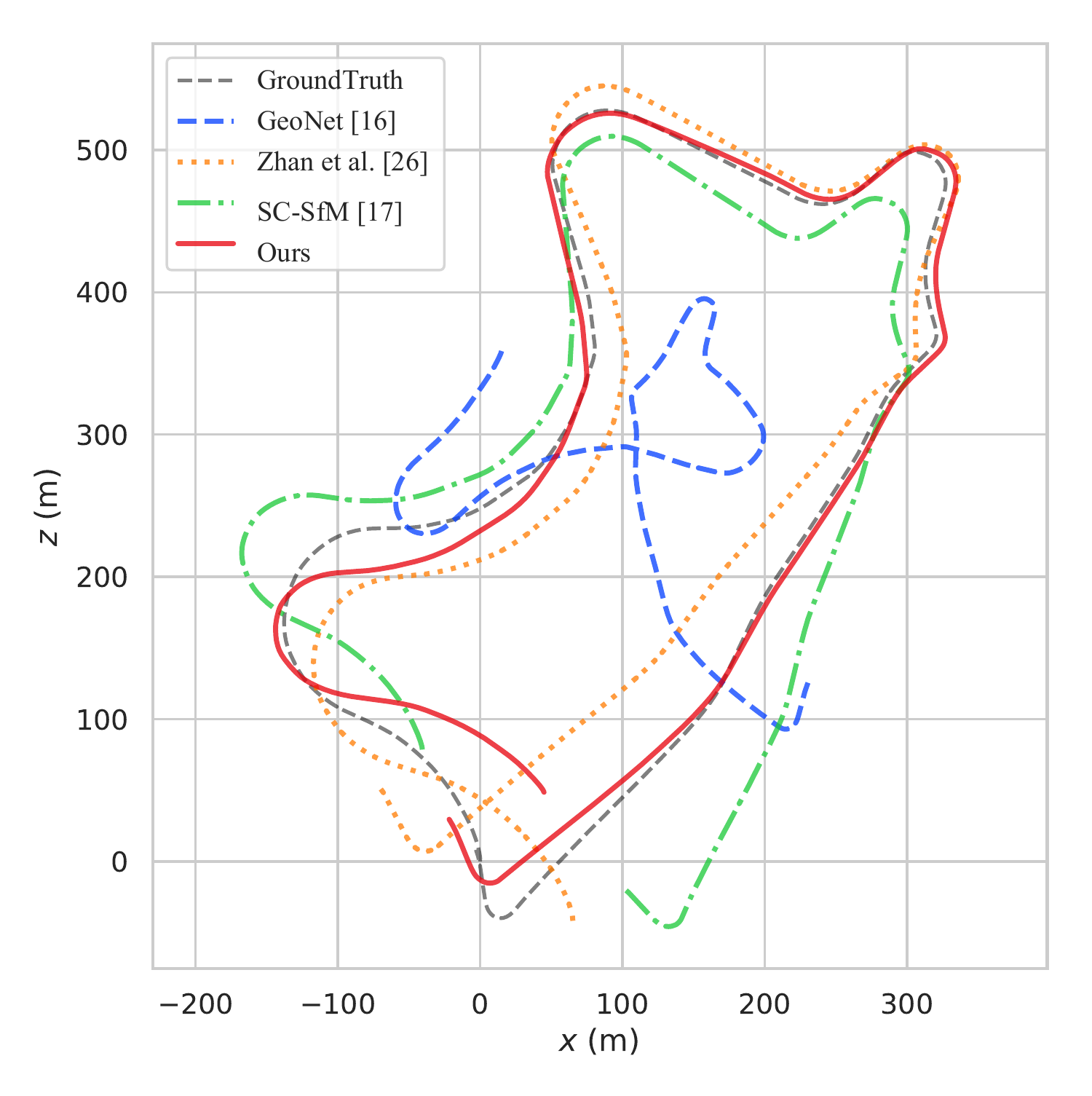}
	}
	\subfigure[Trajectory of Seq. 10 with SE(3) Umeyama alignment]{
		\includegraphics[width = 0.47\columnwidth]{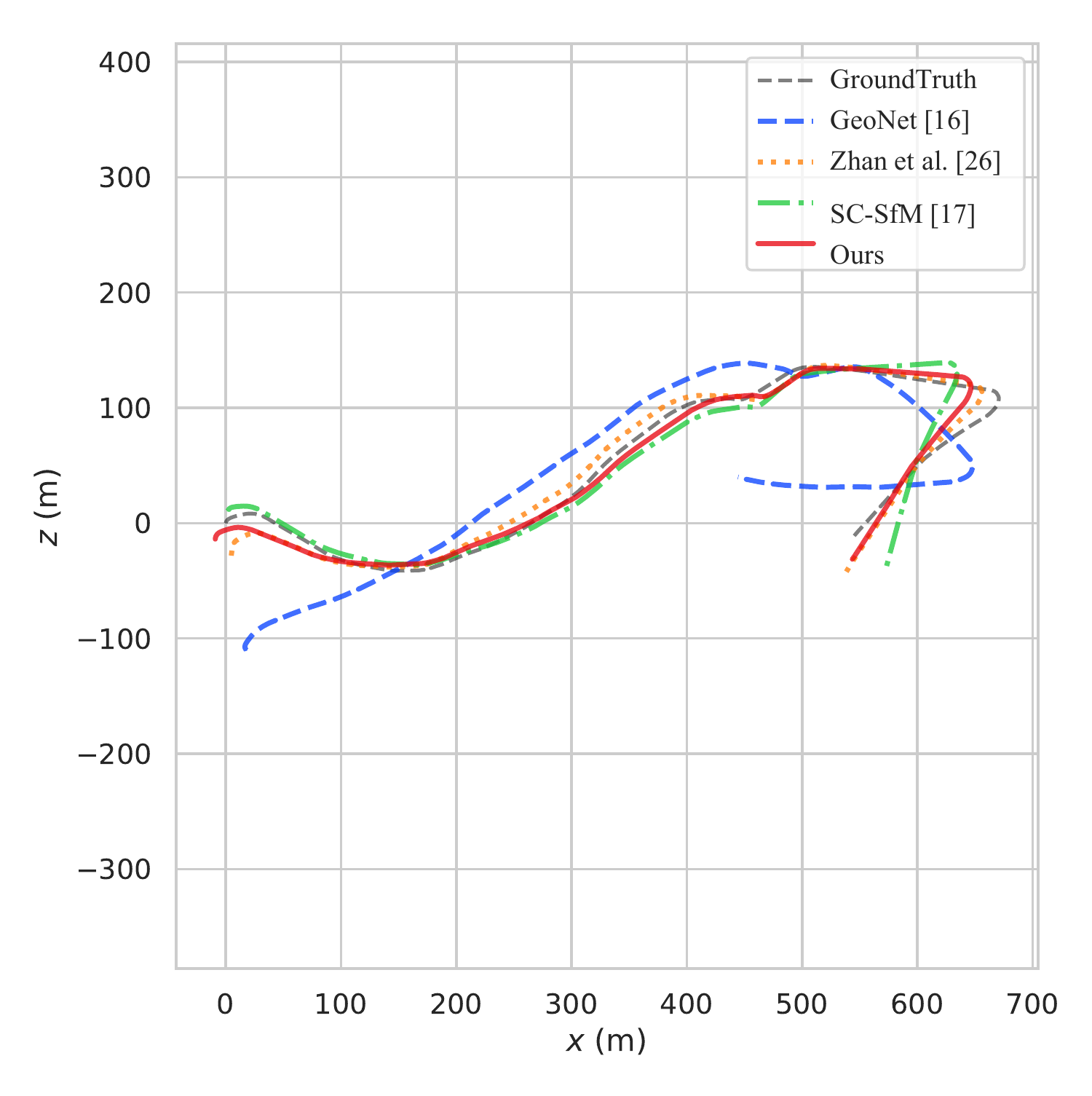}
	}
	
	\caption{Qualitative results on the test sequences of KITTI odometry dataset.}
	\label{fig:fig7}
\end{figure*}

\textbf{Comparison with the methods using GANs:}
We compare our method with the unsupervised monocular methods proposed in \cite{cs2018monocular,almalioglu2019ganvo,li2019sequential}, which introduce adversarial learning into the training framework. Note that the GANVO \cite{almalioglu2019ganvo} generates its depth prediction from a vector, which means that their depth network cannot be used independently and predict the depth in an end-to-end manner. At the same time, the depth network in \cite{li2019sequential} uses a single image as well as the temporal information for the depth estimation, and as a result, this network takes more information than ours on depth estimation. In addition, because of the LSTM module used in their framework in \cite{li2019sequential}, the accuracy of the depth network depends on the image sequences, and the depth network cannot predict the depth map from only a single image, which limits its practical applications. Although our DepthNet is jointly trained with other networks in an unsupervised manner, it can be used independently during testing and has the ability to accurately generate depth maps from single images, which enables it to perform depth estimation on some independent images such as network images. The quantitative results are shown in Table \ref{Tab02}, and our method obtain the competitive results.

\textbf{Comparison with previous work: }
The qualitative and quantitative results of our DepthNet are evaluated by public metrics and are shown in Fig. \ref{fig:fig6} and Table \ref{Tab03}. Comparing the ground truth with our predicted depth maps in Fig. \ref{fig:fig6}, our deep network predicts the depth information of the geometric structures in the scenes, such as trees, streets, cars, and buildings.
Note that higher resolutions include detailed geometric details, and thus, we divide the results of the different methods according to the resolution of their input images for fairness. We choose images with different resolutions as input to our DepthNet.
As shown in Table \ref{Tab03}, we obtain competitive performance on end-to-end monocular depth prediction with different resolution, 416$\times$128 and 640$\times$192. Our results also prove that a high-resolution input is conducive to improvement in the accuracy.

\textbf{Evaluation based on other datasets:}
Previous methods \cite{yin2018geonet,bian2019depth,ranjan2019competitive} have proved that depth networks will get better results if the models are pre-trained on the Cityscapes datasets \cite{Cordts2016Cityscapes} before training on KITTI dataset \cite{geiger2013vision}. Following this approach, we first use the Cityscapes dataset to pre-train our total framework, and then the pre-trained model is further trained by the KITTI training set, which is shown as ``CS+K'' in Tables \ref{Tab03} and \ref{Tab05}.

To further evaluate the performance of our depth network on other data domains, following \cite{xiong2020selfsupervised}, we directly test our depth model trained by ``CS+K'' on the Make3D test set \cite{saxena2008make3d}. As shown in Table \ref{Tab04}, our DepthNet also shows good performance when compare with previous methods, which means that our model has better transferability. Our qualitative results are illustrated in Fig. \ref{fig:fig6}.

\begin{table}[]
	
	\scriptsize
	
	\centering
	
	\caption{\textit{Test on Make3D Dataset \cite{saxena2008make3d}}}
	
	\label{Tab04}
	\resizebox{0.95\columnwidth}{!}{
		\begin{tabular}{c|c|c|c|c}
			
			\toprule
			
			Method  &  Abs Rel	&  Sq Rel   &  RMSE  &  RMSE log \\
			\hline
			SfMLearner \cite{zhou2017unsupervised}  & 0.383 &  5.321 &  10.47  & 0.478	\\
			DDVO 	\cite{wang2018learning}			& 0.387 &  4.720 &  8.09   & 0.204  \\
			SC-SfMLearner \cite{bian2019depth}    	& 0.337 &  3.302 &   7.162  & 0.171 \\
		    Xiong \textit{et al.}  \cite{xiong2020selfsupervised}	& 0.320 &  3.170 &   7.062  & \textbf{0.163} \\
			Ours					& \textbf{0.312} & \textbf{2.814}  & \textbf{6.863} & \textbf{0.163}	 \\
			\bottomrule
			
		\end{tabular}
	}
\end{table}

\subsection{Trajectory Prediction}

For PoseNet, following Zhan \textit{et al.} \cite{zhan2018unsupervised} and Bian \textit{et al.} \cite{bian2019depth}, the standard evaluation tools provided by the dataset are used to evaluate the full predicted trajectories, which are different from previous monocular DL-based pose evaluation methods \cite{zhou2017unsupervised,yin2018geonet,ranjan2019competitive,feng2019sganvo}.
Table \ref{Tab05} shows the average rotation and translation errors of the predicted trajectories on the KITTI odometry sequences 09-10.
As shown in Table \ref{Tab05}, comparing the results of ``Ours(Without $\mathcal{L}_{\textrm{scale}}$)'' and ``Ours'', the proposed scale consistency loss $\mathcal{L}_{\textrm{scale}}$ is effective in constraining the scale consistency in the trajectory prediction.
Although the method in \cite{li2019sequential} also considers the scale-inconsistency of pose estimation, the generated trajectories are neither evaluated in the article nor published online, and thus, we cannot make a quantitative comparison with their trajectories.
The visual results are shown in Fig. \ref{fig:fig7}, which are drawn by evo tools \cite{grupp2017evo} with automatic scale alignment for the full trajectory.
The method \cite{zhan2018unsupervised} trained with stereo image pairs does not have the problem of scale inconsistency because it learns the scale information from stereo image pairs during the training.
Monocular methods \cite{zhou2017unsupervised,yin2018geonet,wang2019recurrent} train their network with monocular sequence and suffer from per-pose scale ambiguity and inconsistency.
At the same time, the scale information between the different snippets predicted by PoseNet is inconsistent in such a way that it cannot provide an accurate trajectory. Bian \textit{et al.} tackle this problem by geometric alignment, and our framework obtains a better result than theirs. As shown in Table. \ref{Tab05}, our model, when trained on ``CS+K'', does not obtain a large improvement similar to others \cite{zou2018df,bian2019depth} when compared with the model trained on ``K''.

\begin{table}[]
	
	\scriptsize
	
	\centering
	
	\caption{\textit{ Visual odometry results on KITTI odometry dataset \cite{geiger2013vision}}}
	
	\label{Tab05}
	\resizebox{0.95\columnwidth}{!}{
		\begin{tabular}{c|c|cc|cc}
			
			\toprule
			\multicolumn{2}{c}{}& \multicolumn{2}{c}{Seq. 09} & \multicolumn{2}{c}{Seq. 10}  \\
			\cmidrule(r){3-4} \cmidrule(r){5-6}
			
			Method  & Supervision	&  $t_{err}(\%)$		&  $r_{err}(^{\circ}/100m)$      &  $t_{err}(\%)$    &   $r_{err}(^{\circ}/100m)$ 	 \\
			\hline
			ORB-SLAM \cite{mur2017orb} & -	&  15.30    	&  0.26  &  3.68   &   0.48  \\
			\hline
			SfMLearner \cite{zhou2017unsupervised} &Mono./ K	&  17.84   	&  6.78  &  37.91   &   17.78 	 \\
			GeoNet 	\cite{yin2018geonet}		&Mono./ K	&  41.47    &  13.14 &  32.74   &   13.12	 \\
			Zhan \textit{et al.} \cite{zhan2018unsupervised} &Stereo./ K	&  11.93    &  3.91  &  12.45   &  3.46  \\
			Bian \textit{et al.}	\cite{bian2019depth} 		&Mono./ K	&  11.2    	&  3.35  &  10.1    &   4.96 	 \\
			Wang \textit{et al.} \cite{wang2019recurrent} 		&Mono./ K	&  9.88   	&  3.40  &  12.24   &  5.20	 \\
            Ours(Without $\mathcal{L}_{scale}$)         	&Mono./ K	& 12.83  & 3.87  &  13.58   &   4.33	 \\
			Ours 								&Mono./ K	& \textbf{8.71}  & \textbf{3.10}  &  \textbf{9.63}& \textbf{3.42}	 \\
			\hline
			Bian \textit{et al.}	\cite{bian2019depth} 		  &Mono./ CS+K	&  8.24    	&  2.19  &  10.7    &   4.58 	 \\
            Xiong \textit{et al.}  \cite{xiong2020selfsupervised} &Mono./ CS+K	& \textbf{5.85}  & \textbf{1.73}  &  10.11   &   3.89	 \\
			Ours 								&Mono./  CS+K	 & 8.13  & 2.64 &  \textbf{9.74} & \textbf{3.58}	 \\

			\bottomrule
			
		\end{tabular}
	}
\end{table}

In addition, as shown in Table \ref{Tab05}, because of the strong back-end optimization, ORB-SLAM \cite{mur2017orb} shows more powerful performance in position and orientation prediction than deep learning-based VO methods.
Although there is still a large gap compared with the traditional VO method \cite{mur2017orb}, our PoseNet has the ability to predict the full trajectory over a long monocular video in an end-to-end manner.


\section{Conclusions}
In this paper, we present an unsupervised monocular depth and VO estimation framework that has scale consistency through adversarial learning methods. The proposed method considers the impact of incomplete reconstruction caused by dynamic objects, occlusions and visual field changes on the reconstruction loss and adversarial loss, which in turn affects the training of the discriminator and generator networks.
This paper tackles these problems by introducing MaskNet. The mask predicted by MaskNet is used to reduce the impact of incomplete reconstruction on the reconstruction loss. In addition, we design a BMP to preprocess real images to produce the same data distribution as the unreconstructed areas on the synthesized images, thus restoring the balance between the generator and discriminator. Furthermore, we tackle the scale-inconsistency in pose and depth estimation by introducing an adaptive constraint. With the proposed adversarial learning framework, our depth model shows competitive results with state-of-the-art methods, and our pose model has the ability to provide a global trajectory over a long monocular sequence, which is meaningful for practical applications. In the future, we will improve the geometric constraints for more accurate trajectory prediction.

\ifCLASSOPTIONcaptionsoff
  \newpage
\fi



%
{
	\bibliographystyle{IEEEtran}
	\bibliography{egbib}
}

%
\newpage




\end{document}